\documentclass[conference]{IEEEtran}
\IEEEoverridecommandlockouts
\usepackage{cite}
\usepackage{amsmath,amssymb,amsfonts}
\usepackage{algorithmic}
\usepackage{graphicx}
\usepackage{textcomp}
\usepackage{xcolor}
\usepackage{caption}
\usepackage{subcaption}
\def\BibTeX{{\rm B\kern-.05em{\sc i\kern-.025em b}\kern-.08em
    T\kern-.1667em\lower.7ex\hbox{E}\kern-.125emX}}
\begin{document}

\title{Mechanistic Interpretability of Reinforcement Learning Agents\\}

\author{\IEEEauthorblockN{Tristan Trim}
\IEEEauthorblockA{\textit{University of Victoria} \\
Victoria BC, Canada \\
tristantrim@gmail.com}
\and
\IEEEauthorblockN{Triston Grayston}
\IEEEauthorblockA{\textit{University of Victoria} \\
Victoria BC, Canada \\
graystontriston@gmail.com}

}

\maketitle

\begin{abstract}
This paper explores the mechanistic interpretability of reinforcement learning (RL) agents through an analysis of a neural network trained on procedural maze environments. By dissecting the network's inner workings, we identified fundamental features like maze walls and pathways, forming the basis of the model's decision-making process. A significant observation was the goal misgeneralization, where the RL agent developed biases towards certain navigation strategies, such as consistently moving towards the top right corner, even in the absence of explicit goals. Using techniques like saliency mapping and feature mapping, we visualized these biases. We furthered this exploration with the development of novel tools for interactively exploring layer activations.

\end{abstract}

\section{Introduction}
Mechanistic interpretability is an emerging field within artificial intelligence research, dedicated to the reverse engineering of neural networks to understand their internal mechanisms after they have been trained to perform specific tasks. This area of study aims to unravel the complexities of how neural networks process information and make decisions, which is often considered a "black box" due to the intricate and opaque nature of these models. By dissecting the inner workings of neural networks, researchers can gain insights into the underlying principles that guide their functioning, potentially leading to more transparent, robust, and trustworthy AI systems. 

This project continues the exploration of the models introduced in “Goal Misgeneralization in Deep Reinforcement Learning" \cite{goal misgen}. The paper explored the effects of training a reinforcement learning agent on an environment modified to exhibit certain behaviours that would not be present in testing to explore what effects that might have on the agent itself. This includes having the goal exclusively exist in a specific area during testing, and having redundant actions. That paper explored the effects this would have on the model via primarily statistical methods. Our goal is to delve into pre-trained models and uncover the strategies and patterns employed to navigate and solve mazes using common methods found in the field of mechanistic interpretability. 

Through this investigation, we hope to contribute to the broader field of AI interpretability by providing a detailed case study of a reinforcement learning model with well studied misgeneralization. Our findings could have significant implications for the development of more interpretable and explainable AI systems, which are essential for applications where transparency and accountability are paramount. 

\section{Related Works}

The study "Leveraging Procedural Generation to Benchmark Reinforcement Learning" \cite{procgen} introduced the Procgen Benchmark, a suite of 16 procedurally generated game environments specifically designed to evaluate reinforcement learning (RL) algorithms. These environments are characterized by their diversity and complexity, which are crucial for testing the generalization and robustness of RL agents. Each game environment in the Procgen Benchmark is created using procedural generation techniques, ensuring that every instance presents a unique scenario. This uniqueness is vital for benchmarking because it prevents overfitting to specific game layouts and promotes the development of RL models that can generalize across varied tasks. The benchmark includes a wide range of game genres and difficulty levels, providing a comprehensive platform for assessing the performance of different RL algorithms under diverse conditions.

Building on the Procgen Benchmark, the research presented in "Goal Misgeneralization in Deep Reinforcement Learning" explored the concept of goal misgeneralization \cite{goal misgen}. Goal misgeneralization occurs when an RL agent, although capable of performing tasks outside its training distribution, ends up pursuing incorrect objectives. This misgeneralization poses significant challenges, as it indicates that the agent has not correctly internalized the intended goals of its tasks. The study extended the Procgen environments to investigate this phenomenon, using them as a testbed to analyze how and why RL agents might misgeneralize their goals. Through systematic experimentation, the researchers demonstrated instances where RL agents, despite showing competence in task execution, failed to align their behavior with the desired goals, highlighting the need for better training methodologies and evaluation metrics to ensure that RL agents not only learn effectively but also generalize appropriately to new situations.

\begin{figure}[htbp]
\includegraphics[scale=1]{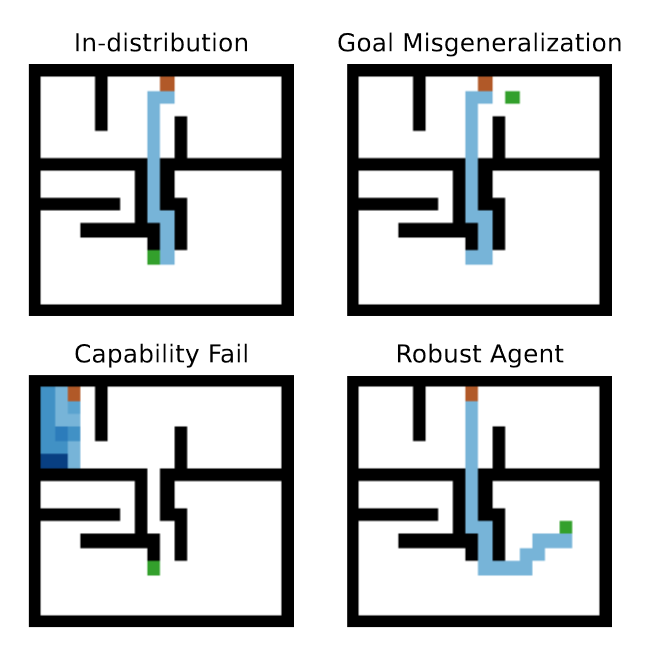}
\caption{Goal Misgeneralization Pic from \cite{goal misgen}. Red is agent, green is the goal}
\label{fig:cheese region maps}
\end{figure}

In the subsequent work "Understanding and Controlling a Maze-Solving Policy Network," researchers selected a partially misgeneralizing policy network trained on the Procgen maze environment for in-depth mechanistic exploration \cite{understanding and controlling}. The maze environment utilized in this study employs Kruskal's algorithm to generate complex maze structures. These mazes are presented as images, with a mouse icon representing the player's position and a cheese icon representing the goal. The study aimed to dissect the internal workings of the policy network, which had shown partial goal misgeneralization, to understand the specific pathways and decision-making processes it used to solve the mazes. By employing techniques such as activation patching, the researchers were able to manipulate specific neuron activations within the network, effectively controlling its behavior. This approach not only provided insights into the network's decision-making process but also demonstrated the potential of mechanistic interpretability techniques to influence and correct the behavior of RL models.

Our project aims to extend the foundational work laid out by these studies by employing a vast array of additional interpretability techniques beyond those already covered. We intend to apply a comprehensive suite of methods, including but not limited to saliency mapping and feature mapping to uncover deeper insights into the inner workings of the Impala network. By leveraging these advanced techniques, we hope to discover new aspects of the network's functionality and decision-making processes that were not previously understood. This extensive application of interpretability methods is intended to provide a more thorough and nuanced understanding of how RL models, like the Impala network, navigate and solve complex tasks such as maze environments.

\section{Experimentation}
This project has us conduct various forms of experimentation on an Impala model pre-trained on a fairly simple environment of procedural generated mazes.

\subsection{Feature Mapping}\label{AA}

The introduction of convolutional networks marked a significant milestone in machine learning, especially for image processing. This innovation enabled models to effectively understand spatial relationships between pixels within localised regions of an image. Additionally, convolutional networks offer greater interpretability compared to traditional dense layers, as their outputs are transformed versions of the input images. These visualisations depict kernel activation's, allowing us to understand the specific features each kernel captures from the input image upon a visual inspection. 

The model that we’re working with has 15 convolutional layers comprising the majority of the model's architecture. We used this as a very preliminary exploration into the model, trying to find redundancies and notable patterns that may arise from the model itself. 

This section works with a 512x512x3 image which is then pre-processed into a 64x64x3 grid, and transposed as it goes into the model. These are shown in Fig 2, and we use them as a ground truth for the rest of this section.

\begin{figure}[!tbp]
  \begin{subfigure}[b]{0.2\textwidth}
    \includegraphics[width=\textwidth]{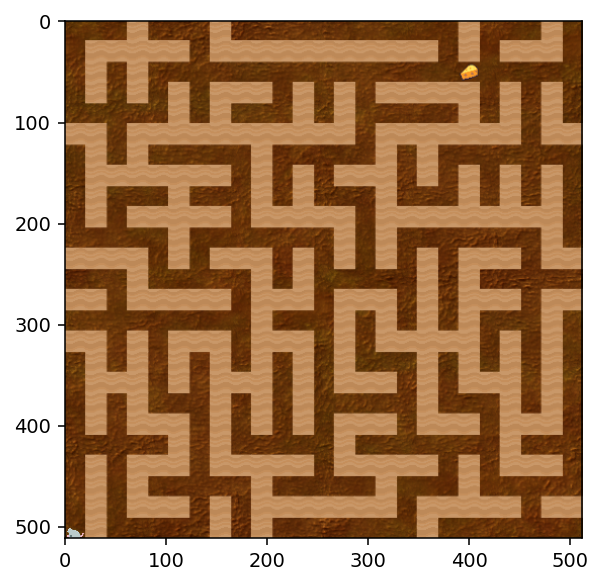}
    \caption{Upscaled Maze}
    \label{fig:upscaled maze}
  \end{subfigure}
  \hfill
  \begin{subfigure}[b]{0.23\textwidth}
    \includegraphics[width=\textwidth]{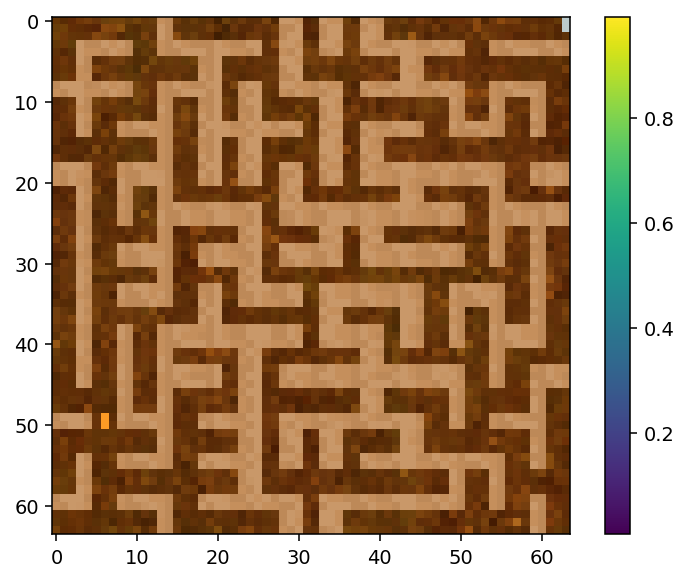}
    \caption{Pre-Processed Maze}
    \label{fig:preprocessed maze}
  \end{subfigure}
  \caption{Mazes}
\end{figure}

Our initial exploratory work focuses on the first layer of the neural network, a crucial step in understanding the foundational patterns recognized by the model. In the context of our task, which involves navigating a maze, we hypothesize that this initial layer would capture the most elementary features present in the environment. These features likely include the walls delineating the boundaries of the maze, the pathways allowing navigation, the location of the cheese (as a reward signal), and the positions of the mice (agents).

\begin{figure}[htbp]
\includegraphics[scale=0.20]{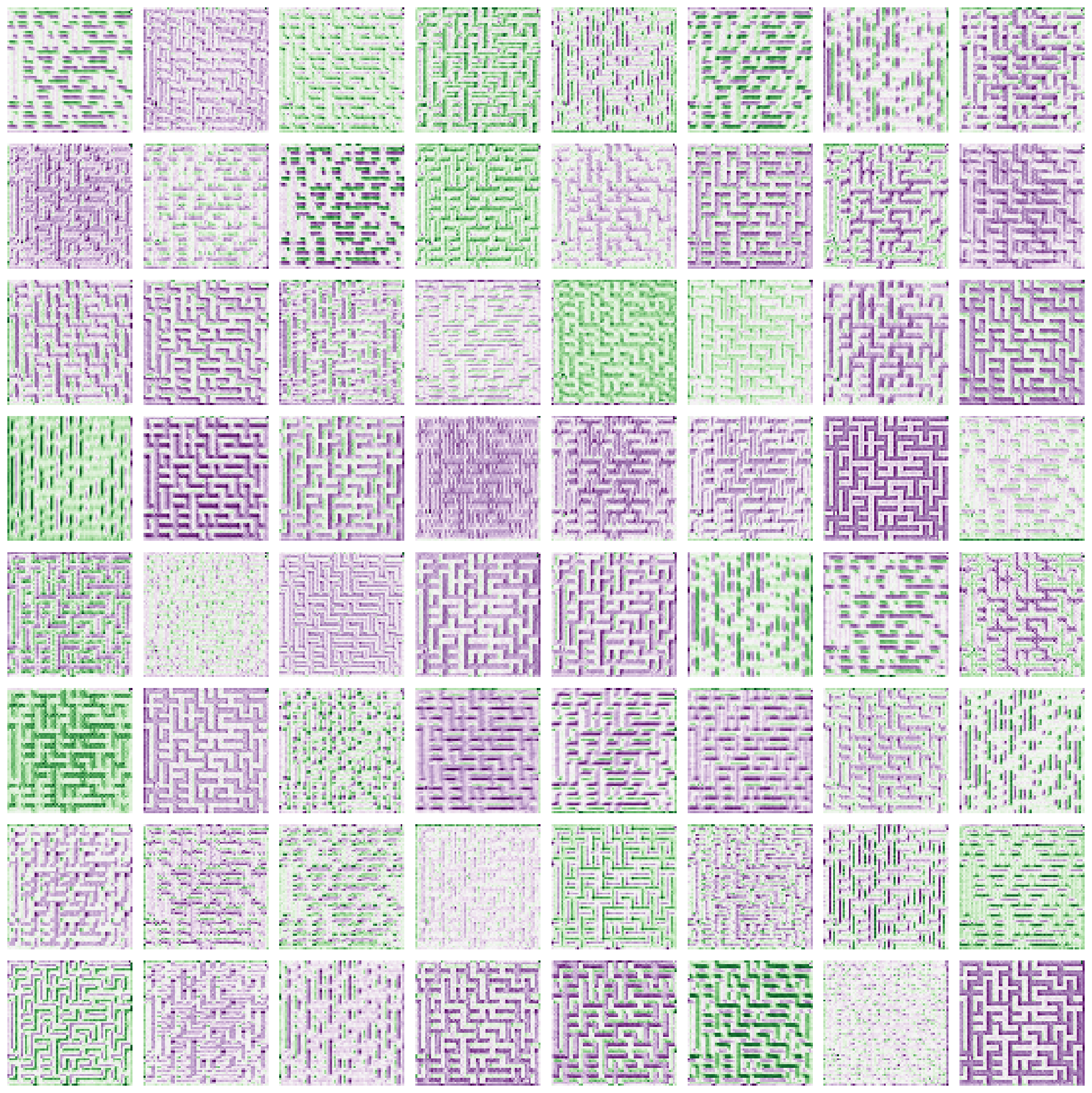}
\caption{First Convolutional Layer Activations. Green = Higher Activations.}
\label{fig:first conv activations}
\end{figure}

To confirm our hypothesis, we analyze the activations of the neurons in the first layer. Activations refer to the responses of neurons to specific inputs. By visualizing these activations, we can infer what features the neurons are responsive to. Our results indicate that certain neurons are indeed selectively responsive to the walls and general structure of the maze. This is evident from the activation maps, which highlight the presence of linear and angular patterns corresponding to the walls and pathways. These are shown in Fig~\ref{fig:first conv activations}.

As a general rule, regions with higher activation levels are represented by the color green in our visualization. We have chosen to employ a diverging colormap rather than a sequential one, as it more effectively highlights both the high and low activation areas. This choice facilitates the differentiation between various levels of neural activity, thereby making it easier to identify significant patterns within the data. A diverging colormap is particularly advantageous for detecting subtle variations in activation that might otherwise be overlooked with a sequential colormap.

Interpolating the data at the first convolutional layer turns out to be a simple task. Activation's are found pretty exclusively at either walls, paths, or corners. What may be interesting to point out, is that at this level there are no obvious kernels that light up specifically around the mouse (agent) or the cheese (goal) despite those being unique pixels highlighting some sort of functionality.  

That being said, the true power of convolutional neural networks lies in the intermediate layers, where patterns detected by lower-level kernels are combined and refined. This process results in more sophisticated and abstract representations, significantly enhancing the model's ability to comprehend and represent the underlying data. 

The next layer we look at is the convolutional layer hidden in the next impala block, referenced in \cite{fig:cheese region maps}. The impala blocks themselves contain pooling and res-net layers, and as a result of this processing the resulting activation maps become significantly harder to interpret visually. We can still, discern several noteworthy observations.

It appears that, in contrast to the earlier layer, the neural network begins to distinctly identify the mouse (actor) within the environment, shown in Figure 4. 

\begin{figure}[htbp]
\includegraphics[scale=0.25]{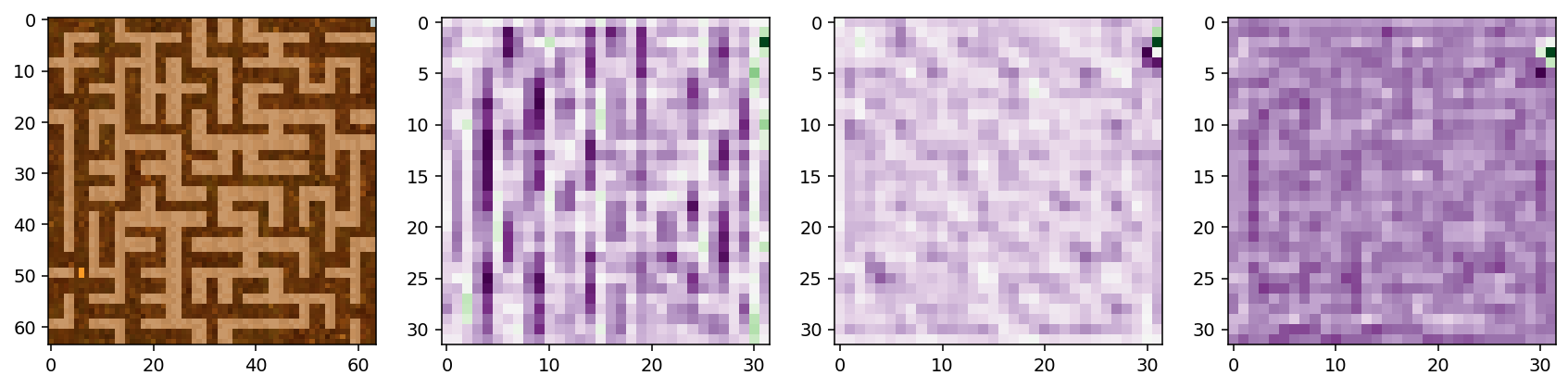}
\caption{Feature Maps Identifying Region Around Cheese}
\label{fig:cheese region maps}
\end{figure}

Several feature maps exhibit notable activations in the vicinity of the top right corner, which corresponds to the mouse's location. Interestingly, however, the peak activations are not situated precisely at the top right pixel, where the mouse is positioned in the environment. Instead, these peak activations seem to be slightly displaced downward, occurring just below the actual position of the mouse. Whether this shift is attributable to the decreased resolution of the feature maps or another underlying factor remains unclear.

It is particularly noteworthy that no feature maps exhibit a similar focused activation pattern for the cheese, which serves as the goal in the environment. This lack of recognition for the cheese is intriguing, as we observe no singular feature map that is activated exclusively at the cheese's location. This poses an interesting question: why would arguably the most unique point on the map be effectively ignored by the model at this level? We conducted several reasons for why this might be the case. The most immediate hypothesis is this being a direct result of goal misgeneralization. That is to say, the cheese in training may be ignored entirely, as it is not a feature necessary for prediction due to it's similar spawning locations. Another hypothesis is that the resolution at this level might obscure the features necessary to recognize the cheese. The network might require a combination of higher resolution and more context-specific information, which is typically processed in deeper layers, to identify the goal accurately. This seems to support previous works exploration on channel 55 \cite{understanding and controlling}.

As we delve deeper into the neural network, the subsequent layers become increasingly complex and challenging to interpret visually. The intricacy of the feature maps in these deeper layers necessitates the adoption of more sophisticated analytical techniques to understand the network’s behavior and decision-making processes.

To streamline the exploration of these deeper layers, it is essential to first comprehend the underlying mechanisms by which mazes are generated and structured.

\subsection{Systemic Maze Investigation.}

"Leveraging Procedural Generation to Benchmark Reinforcement Learning" \cite{procgen} introduces the Procgen Benchmark, a suite of 16 procedurally generated environments designed to evaluate the sample efficiency and generalization capabilities of reinforcement learning (RL) agents. The core motivation behind Procgen Benchmark is to address the limitations of traditional benchmarks, such as the Arcade Learning Environment (ALE), which can lead to overfitting due to their static and repetitive nature. By employing procedural content generation, Procgen Benchmark ensures a near-infinite supply of highly varied and randomized environments, challenging RL agents to develop robust policies that generalize well across different scenarios.

In this project, we utilize procedurally generated mazes within a Python virtual environment (venv) to create a dynamic and challenging setting for training neural networks. The mazes are designed to have one clear path from the agent (mouse) to the goal (cheese), ensuring a consistent objective across different trials. The maze generation process employs various algorithms that ensure a single, solvable path, enhancing the training efficiency and consistency of the reinforcement learning models. These mazes are first created at a high resolution to capture intricate details of the environment.

Once the mazes are generated in some grid format, their high-resolution images are processed by upscaling to enhance visibility and detail. The mazes are then downscaled to a 64x64x3 grid, standardizing the input size for the neural network. This downscaling process reduces the image complexity while preserving essential features, such as walls and paths. 

To introduce variability and prevent overfitting, different seeds are used to generate mazes of varying sizes and configurations. Each seed produces a unique maze, ensuring that the neural network is exposed to a diverse set of environments. This diversity is crucial for developing a model that can generalize well to new, unseen mazes. Additionally, the project includes an interactive tool that allows users to custom-make mazes of a specified size, as shown in \ref{fig:cheese region maps}

\begin{figure}[htbp]
\centering
\includegraphics[scale=0.75]{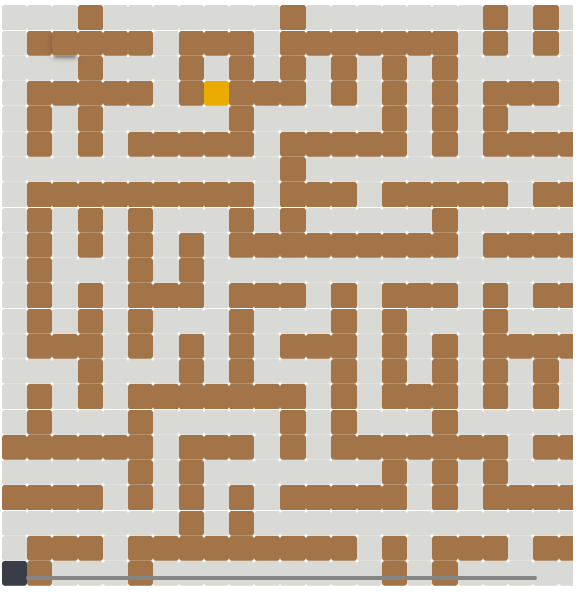}
\caption{Interactive Maze Plotter. (Yellow is cheese, black is mouse)}
\label{fig:cheese region maps}
\end{figure}

This tool provides a hands-on approach to designing maze environments, offering flexibility and control over the training scenarios. Users can create mazes tailored to specific research needs or difficulty levels, further enriching the training dataset.

\subsection{Saliency Mapping}

Saliency mapping is a technique used in neural networks to highlight the regions of an input image that contribute most to the model's prediction. By generating a heatmap over the input, saliency maps indicate which pixels or features the model focuses on, providing insights into the model's decision-making process. This visualization is particularly useful for interpreting and debugging models, as it helps identify potential biases, understand feature importance, and ensure that the model is learning relevant patterns rather than spurious correlations.

Our goal with saliency mapping is to identify specific navigational behaviors of the neural network. We hypothesize that due to the methods of training, the network inherently prefers navigating to the top right corner of the maze, even in the absence of the cheese. This would indicate if the model has developed a bias or default strategy favoring that corner as an optimal path. Removing the cheese from the map and feeding it back into the model, we can see that given complete freedom of movement, the mouse will still prioritize movement to the top-right corner. Given that the model was trained with the PPO algorithm \cite{ppo} we can extract exact probabilities of how likely the model is to make certain moves. 

It is important to note, this model was trained with redundancies on purpose to increase the complexity of generalization. One of the ways in which this was enforced was in a redundant moveset. The agent has 5 effective actions, being all non-diagonal directions and an action that does nothing. The model however, has distinctly 15 different outputs. The largest redundancy is having 7 distinct outputs that do nothing. This is an important note from here on out, as the PPO architecture creates a probability distribution over all of these actions despite redundancies in those actions. Saliency mappings have been manufactured for both singular model outcomes (1 of 15) as well as overall action probabilities (A weighted sum of saliency mapping for the action of moving in a certain direction) for particularly notable examples.

\begin{figure}[!tbp]
  \begin{subfigure}[b]{0.2\textwidth}
    \includegraphics[width=\textwidth]{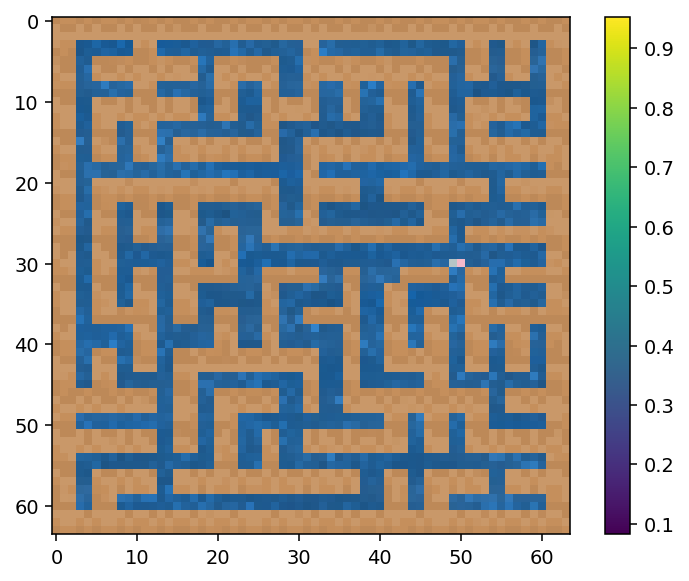}
    \caption{Maze With No Cheese}
    \label{fig:upscaled maze}
  \end{subfigure}
  \hfill
  \begin{subfigure}[b]{0.2\textwidth}
    \includegraphics[width=\textwidth]{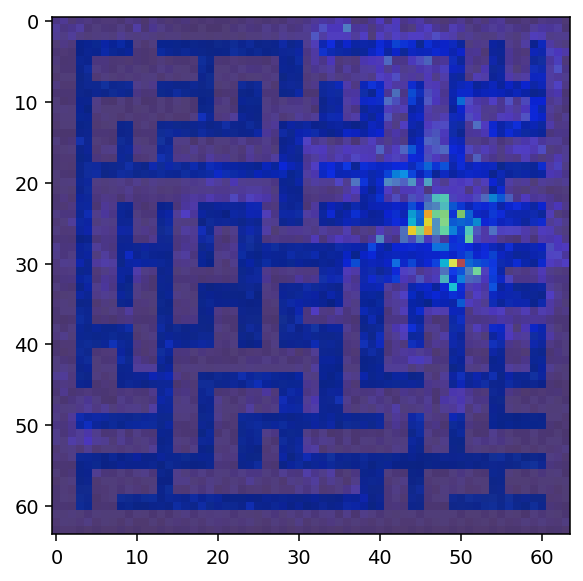}
    \caption{Saliency Map of Maze}
    \label{fig:preprocessed maze}
  \end{subfigure}
  \caption{Saliency Mapping of a Maze with No Goal (Cheese)}
\end{figure}

In the scenario depicted in figure 6, the model exhibits a tendency to move 'UP' with a probability exceeding 80\%, even in the absence of cheese. This behavior suggests that the model has developed a preference for moving upwards, which would be consistent in the agent striving towards the top right corner. This dawns the question, if these training mazes often featured successful outcomes associated with moving 'UP', the model could have encoded this direction as a favorable action, regardless of if that is moving towards the top right corner. We then can test if the actions of moving up, or moving to the top right corner are preferable. We test this with the map shown in Figure 7, where the model can either move up, a move it might find favourable, or left, a move that achieves a step closer to the goal. Indeed, it decides to go left with 96\% probability.

\begin{figure}
    \centering
    \includegraphics[width=.5\linewidth]{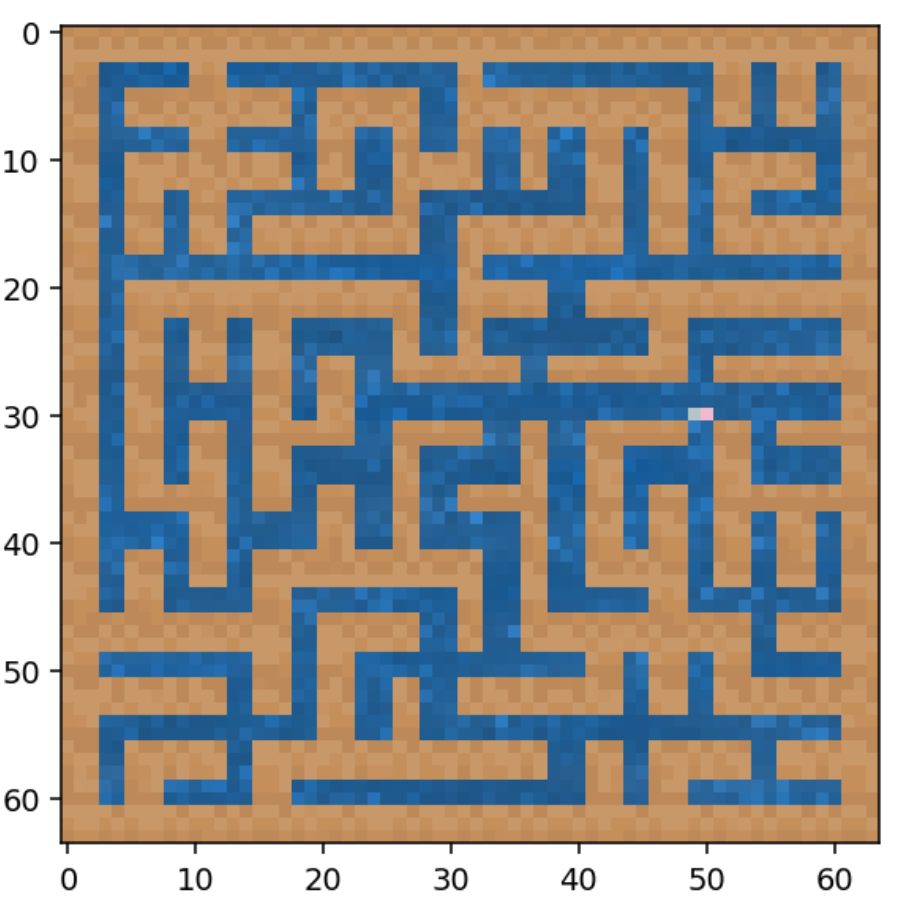}
    \caption{Map Testing Action Probabilities}
    \label{fig:input-dist}
\end{figure}

%secondly, we seek asses
Secondly, we seek to assess whether the network can be directed to move towards the top right corner even when the cheese, the primary goal, is located in the opposite direction. It will indicate the network's ability to prioritize alternative objectives over the default reward-based goal. 

Initial tests involve placing the cheese in a region of the map distant from the top right corner to evaluate the model's perception of its objective. If the model prioritizes the cheese in this scenario, it would suggest minimal misgeneralization, indicating that the model correctly identifies and pursues the primary goal. Conversely, if the model prioritizes the top right corner instead, it would indicate a significant misgeneralization, revealing a potential bias in its learned navigation strategy.

\begin{figure}[!tbp]
  \begin{subfigure}[b]{0.2\textwidth}
    \includegraphics[width=\textwidth]{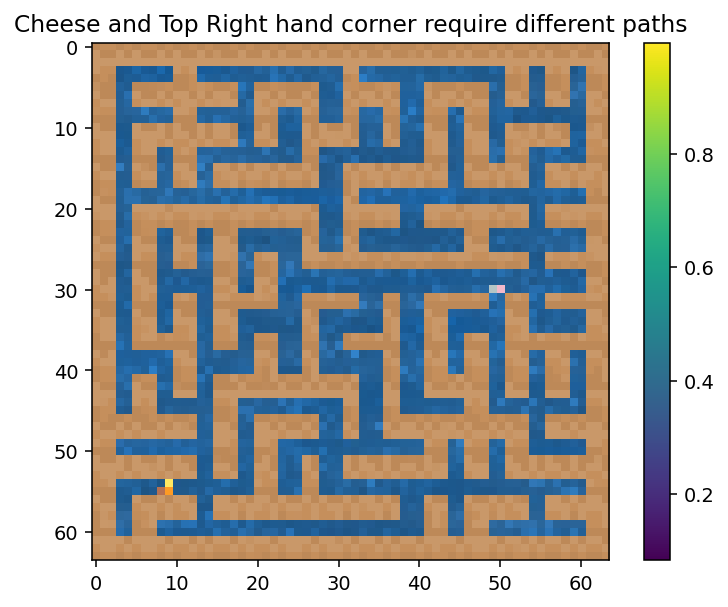}
    \caption{Maze With Cheese in Bottom Left}
    \label{fig:upscaled maze}
  \end{subfigure}
  \hfill
  \begin{subfigure}[b]{0.2\textwidth}
    \includegraphics[width=\textwidth]{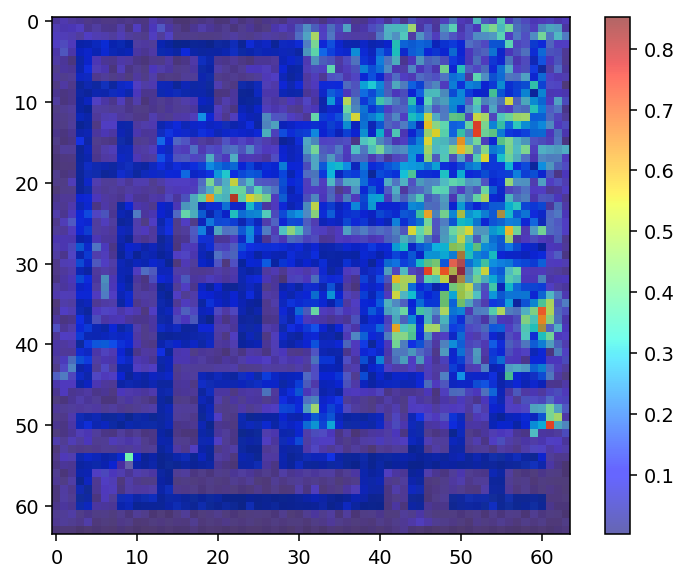}
    \caption{Saliency Map of Maze}
    \label{fig:preprocessed maze}
  \end{subfigure}
  \caption{Saliency Mapping with Cheese in Bottom Left}
\end{figure}

Figure 8 shows that the model heavily highlights the top right region of the map, noting both the mouse's current location and a broad area in the top right corner. The agent predicts it will move 'RIGHT' with a 69\% probability. Interestingly, the agent does recognize the cheese when making this decision, but to a minimal extent. This observation led us to question whether the model perceives both regions as 'goals' and if it uses an intrinsic metric to compare the time steps needed to reach these goals.

One might start to question the model's process of reasoning. The prioritization of the top right hand corner might be due to the large distance away it is from the cheese. Further, there exists a possibility that the top right corner serves as a subgoal for the cheese is intriguing because it suggests a hierarchical approach in the model's navigation strategy. This would imply that the model might be using intermediate targets to simplify the pathfinding task, breaking down the journey into more manageable segments.

We can test both of these hypothesise with figures 9 and 10. 

Figure 9 illustrates an intriguing behavior of the model, where despite high activation near the cheese, the agent initially moves in the opposite direction towards the top right corner. The observed pattern implies that the presence of the cheese in the environment influences the agent’s movement towards the top right. This result, at first, is quite un-intuitive. A high activation of the goal resulting in an action moving in the exact opposite direction may hint at a model's inability to orient itself. Yet, that consideration does not stay consistent with the agent itself, as previous studies show the mouse can travel to the top right hand corners consistently with no orientation problems causing traversal into dead ends. This caused us to consider the possibility of hierarchical sub goals. This decision to move right instead of directly towards the cheese, which shows minimal activation via the saliency maps, suggests that the model might treat the top right corner as a sub goal. 

If our agent has established the top right corner as a hierarchical subgoal, one would expect it to transition towards the cheese after reaching that subgoal. We can test this theory using Figure 10, which shows the agent positioned in the top right corner. Surprisingly, even in this position, the agent does not follow the straightforward path to the cheese, and instead decides to move down with a probability of roughly 51\%. Furthermore, it has a combined probability of less than 10\% of moving left, towards the cheese.

\begin{figure}[!tbp]
  \begin{subfigure}[b]{0.2\textwidth}
    \includegraphics[width=\textwidth]{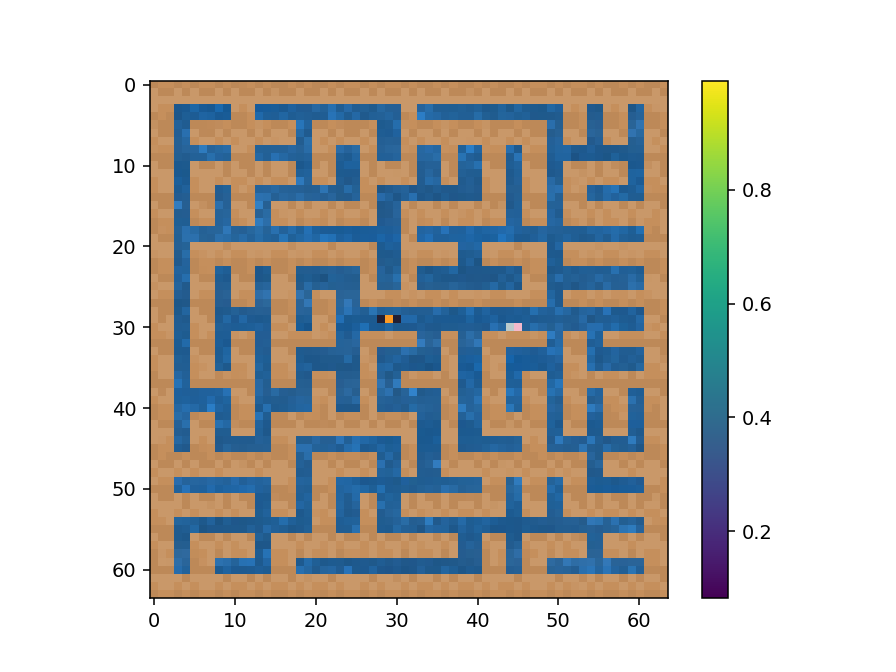}
    \caption{Cheese Equidistant from Top Right}
    \label{fig:upscaled maze}
  \end{subfigure}
  \hfill
  \begin{subfigure}[b]{0.2\textwidth}
    \includegraphics[width=\textwidth]{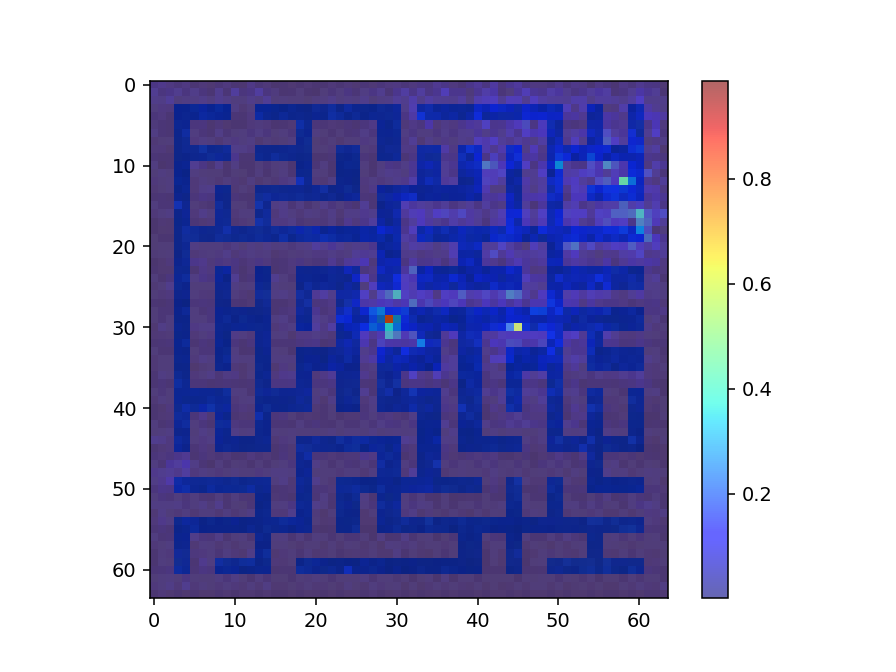}
    \caption{Saliency of Equidistant Map}
    \label{fig:preprocessed maze}
  \end{subfigure}
  \caption{Saliency Map of Mouse ~Equidistant from Cheese and Top Right Boundary}
\end{figure}

\begin{figure}[!tbp]
  \begin{subfigure}[b]{0.23\textwidth}
    \includegraphics[width=\textwidth]{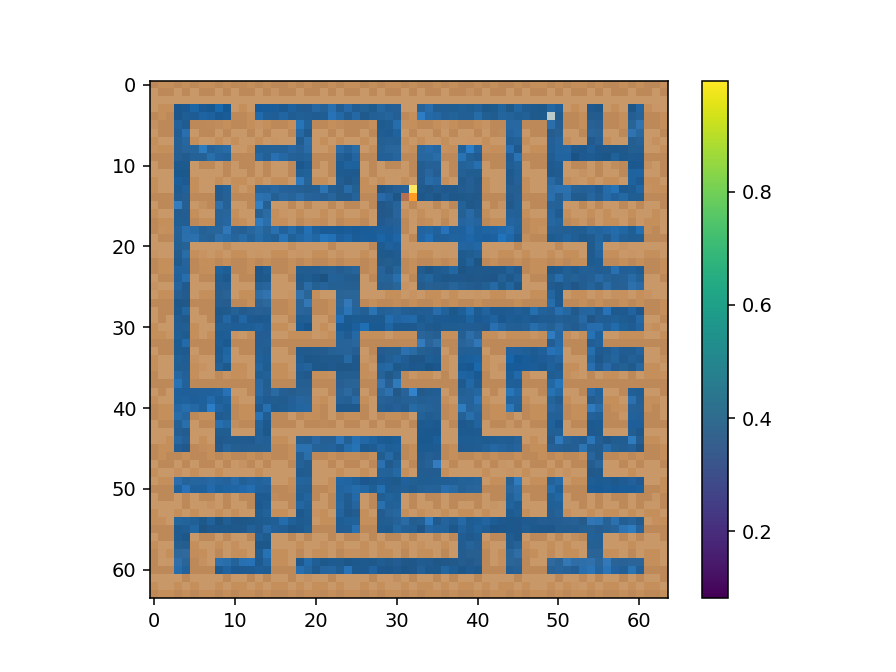}
    \caption{Cheese Bottom Left}
    \label{fig:upscaled maze}
  \end{subfigure}
  \hfill
  \begin{subfigure}[b]{0.23\textwidth}
    \includegraphics[width=\textwidth]{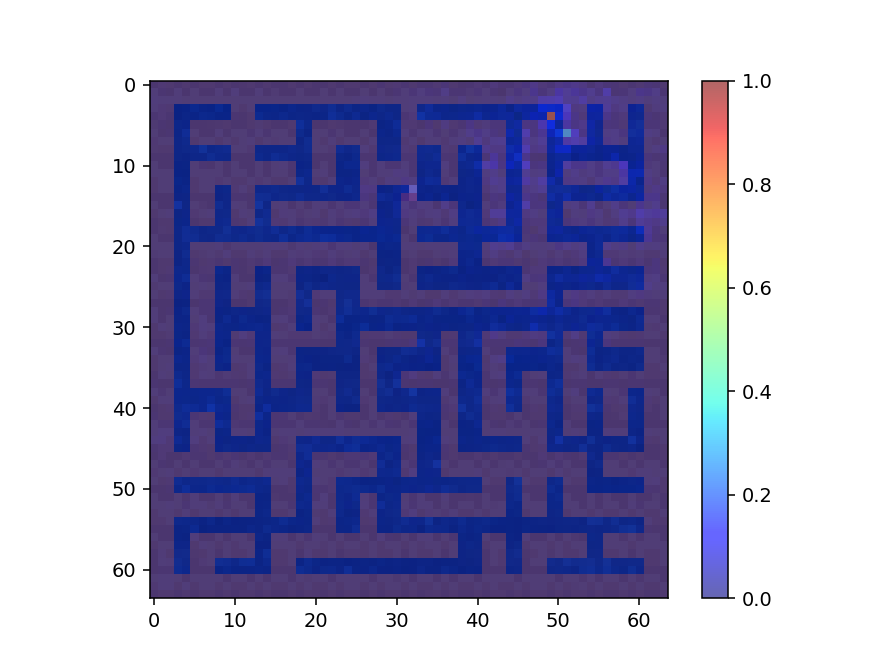}
    \caption{Saliency Map}
    \label{fig:preprocessed maze}
  \end{subfigure}
  \caption{Lack of Activation to Cheese after being in Top Right Corner}
\end{figure}

\subsection{Interactive Distribution Coloring}

\begin{figure}
    \centering
    \includegraphics[width=1\linewidth]{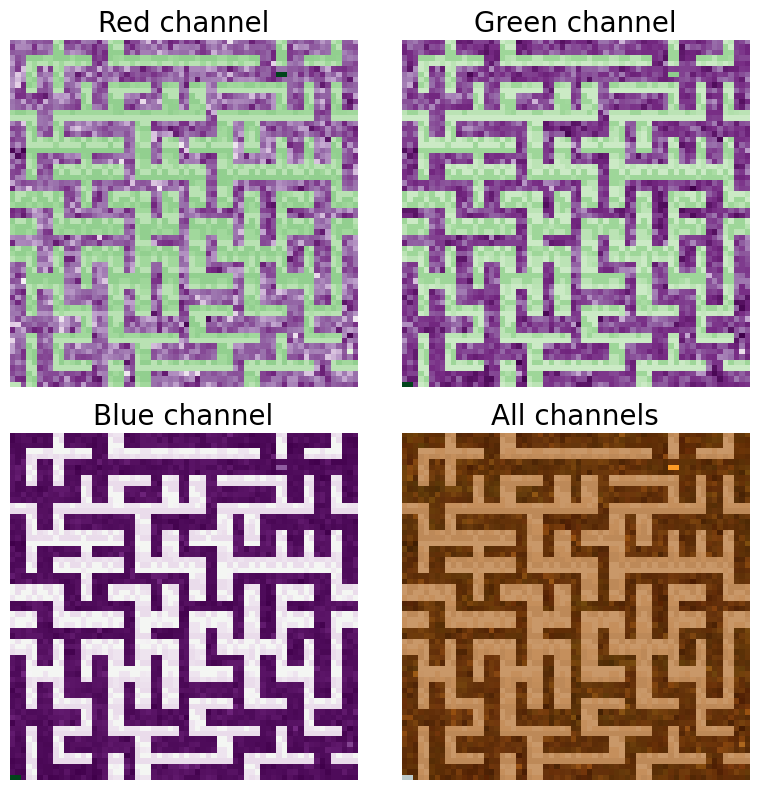}
    \caption{The input image is shown separated into individual channels as is done in feature mapping, and is shown with all channels together. Notice that with all channels together the 4 classes of pixels (block, floor, mouse, and cheese) can easily be seen. To distinguish them in the other images requires a comparison of the pixel values between images. In general, you don't know how close two pixels are to the same color until you have compared their values across all channels.}
    \label{fig:block metaphor}
\end{figure}

Feature mapping has the advantage that it very directly shows you what the neural network is doing, but it has a shortcoming which can be explained by analogy. In block printing, multiple blocks are carved to be used as stamps, each adding their own color to the image that is printed. Fig~\ref{fig:block metaphor} shows what these blocks might look like for the input image to our policy network. Each convolutional layer transforms the image, taking it from 3 channels to 64 and then to 128, but it remains an image that could be printed on a single image if we could see 64 color channels rather than only 3. The key insight is to notice about the input that there are really only 4 colors, not $\mathbb{R}^3$. Or rather, The pixel values cluster around 4 locations in red-green-blue-space: dark brown, light brown, yellow, and gray. If the same thing is true of the images produced by the 64 "printing blocks" of feature mapping, then we could identify the clusters and color pixels based on which cluster they are in. What this looks like for the input image can be seen in Fig~\ref{fig:input-dist}.

%% xxx images for interactive distribut.

\begin{figure}
    \centering
    \includegraphics[width=.5\linewidth]{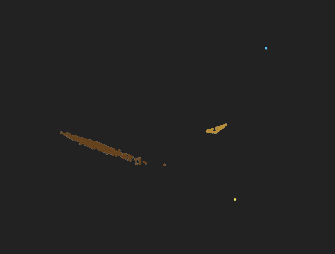}
    \caption{The distribution of pixel colors found in the input. The dark brown floor varies the most. Both the cheese and mouse appear to be a single color, and thereby occupying a single point in red-green-blue-space.}
    \label{fig:input-dist}
\end{figure}

To explore this concept, we have developed a methodology involving 2 novel tools we have developed: PixCol, an interactive pixel class coloring tool, and NDSP, an n-dimensional scatter plot and distribution classification explorer. As an overview, the method involves: 1) using a clustering method in color space to classify the pixels, 2) using PixCol to color and label each pixel class, 3) using NDSP for viewing and editing the classification in the pixel distribution. Finally, that classification and coloring can be used as a reference for iterating the process on the next layer of the neural net.

\subsubsection{Pixel Clustering}

The output of any conv layer will be a tensor with shape = (number of dimensions, pixel width, pixel height). To cluster the pixels, the width and height are flattened together, and each pixel is treated as an independent datum living in $\mathbb{R}^{\text{\#}channels}$. The pixels can then be classified by any clustering method. We explored using hierarchical clustering, k-means, and manually clustering the points with NDSP.

Each of these methods seemed to have it's advantages. At the shallow layers, hierarchical clustering was fast and very accurate at giving clusters with meaningful semantics, while k-means tended to cut clusters, putting pixels into classes with other pixels seeming to represent dissimilar things. In both cases manual classification with NDSP took a great deal of effort, and gave better results with enough effort, however due to the nature of working with $\mathbb{R}^{64}$, human error often introduced different kinds of errors than seen in the clustering algorithms. Because of this, the best approach seems to be classifying with a computer algorithm and then manually editing the classification with NDSP to remove noticeable misclassification.

\subsubsection{Pixel Coloring}

Once we have a classification of pixels, we can use PixCol to visualize it. PixCol is an interactive plot created in jupyterlab with matplotlib that lets you easily create a colored image based on the information in the pixel classification and in the previous layers.

\begin{figure}[!tbp]

    \begin{subfigure}[b]{0.23\textwidth}
      \includegraphics[width=\textwidth]{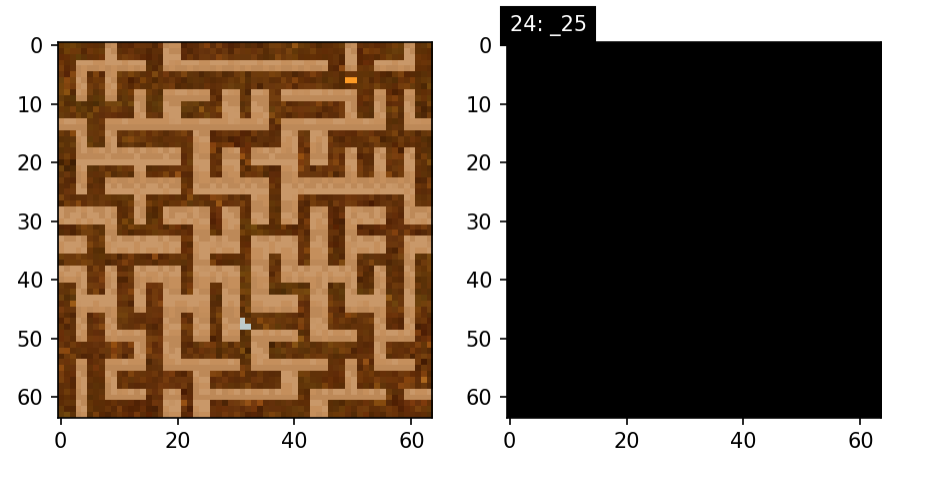}
      \caption{PixCol Initial View}
      \label{fig:pxcl-init}
    \end{subfigure}
    \hfill
    \begin{subfigure}[b]{0.23\textwidth}
      \includegraphics[width=\textwidth]{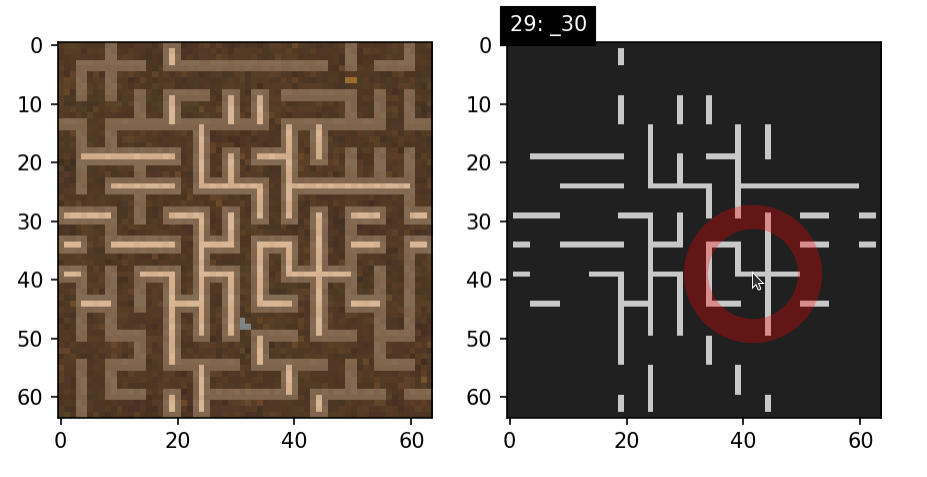}
      \caption{Result of Mouseover}
      \label{fig:pxcl-mouseover}
    \end{subfigure}

    \begin{subfigure}[b]{0.23\textwidth}
      \includegraphics[width=\textwidth]{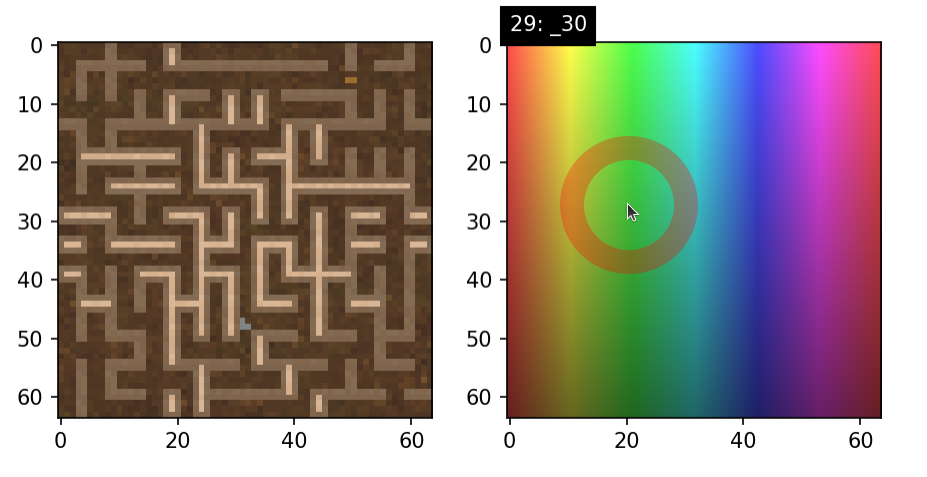}
      \caption{Choosing a Color}
      \label{fig:pxcl-select-color}
    \end{subfigure}
    \hfill
    \begin{subfigure}[b]{0.23\textwidth}
      \includegraphics[width=\textwidth]{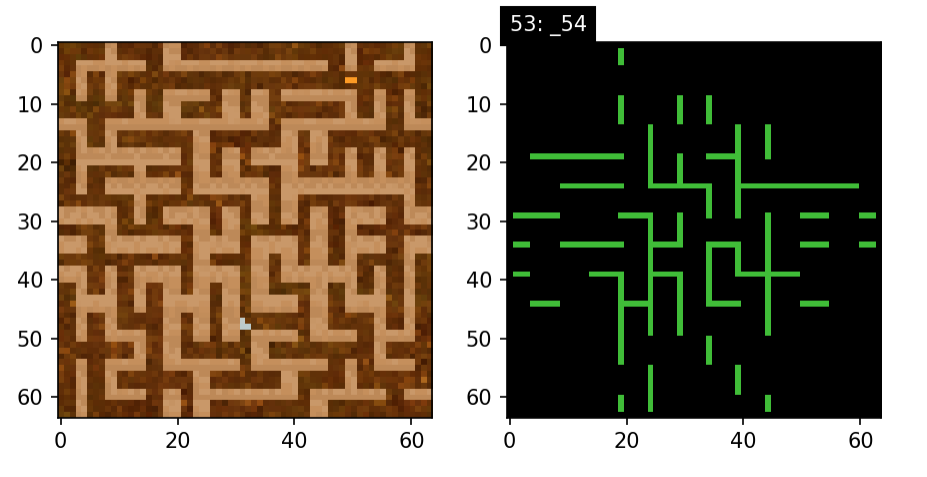}
      \caption{Result of Color Choice}
      \label{fig:pxcl-colored}
    \end{subfigure}
    
    \begin{subfigure}[b]{0.23\textwidth}
      \includegraphics[width=\textwidth]{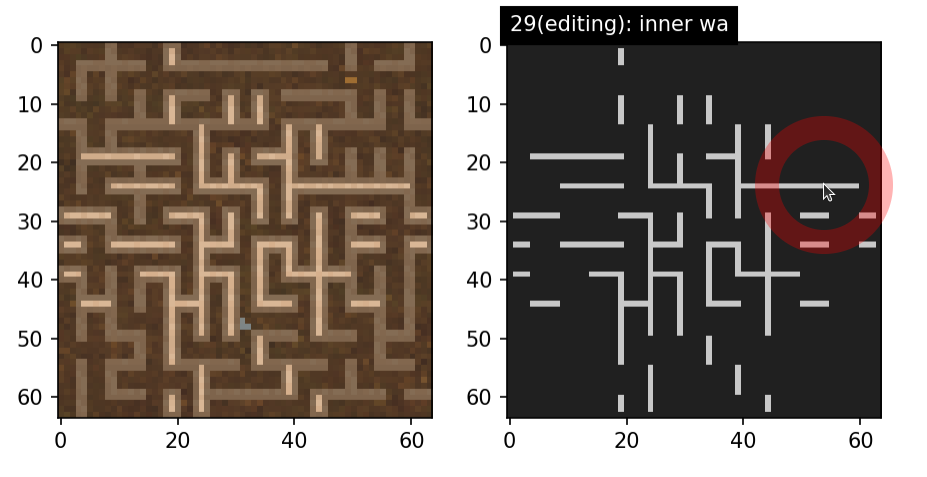}
      \caption{Typing Label}
      \label{fig:pxcl-typing}
    \end{subfigure}
    \hfill
    \begin{subfigure}[b]{0.23\textwidth}
      \includegraphics[width=\textwidth]{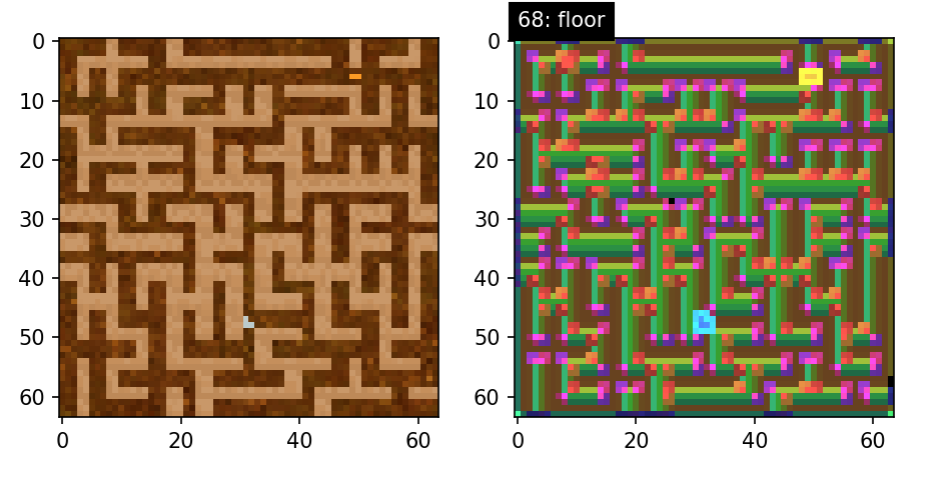}
      \caption{Fully Colored}
      \label{fig:pxcl-fully-colored}
    \end{subfigure}
  
  %%\hfill
  \caption{PixCol usage.    
}
  \label{fig:pxcl-usage}
\end{figure}

Fig~\ref{fig:pxcl-init} shows the view after initialization. On the left, the reference layer is shown. This can be the previous layer, an earlier layer, or any other image that may be useful to cross reference. On the right is the view of the current classification image. Since it has just been initialized, all the pixel classes have been assigned black, and the whole image appears black. When the user hovers their mouse over a pixel, all of the pixels in the same class are highlighted in white, as can be seen in Fig~\ref{fig:pxcl-mouseover}. The user may click on the class to select a color for it. Selecting a color and the resulting change to the image is shown in Fig~\ref{fig:pxcl-select-color} and Fig~\ref{fig:pxcl-colored}. While hovering over a class, the user may type in a label and press enter to assign it to the pixel class, Fig~\ref{fig:pxcl-typing}. Finally, Fig~\ref{fig:pxcl-fully-colored} shows the results of fully coloring the classification for the output of the block1.conv layer. These results are discussed in more depth in (4).

\subsubsection{Pixel Distribution}

Having put a color to every class and understood what conditions in the previous layer cause them may give the impression that we have the full picture. However, these are flat colors. The clusters they represent are not flat. They may be closer or further from any other nearby cluster, or they may have been misclassified. This motivates us to look at the clusters themselves. PCA, T-SNE, UMAP, or other more common statistical methods could be applied, however the interactivity of our approach fits well with "Visualizing Neural Networks with the Grand Tour" \cite{grand-tour}, and our project to extend it, started as a class project \cite{tt-seng310-proj}. We have extended this into a tool we are calling n-dimensional scatter plot (NDSP), which is very helpful in making sense of and editing classifications in pixel distributions.

\begin{figure}[!tbp]

     \centering
    \begin{subfigure}[b]{0.36\textwidth}
      \includegraphics[width=\textwidth]{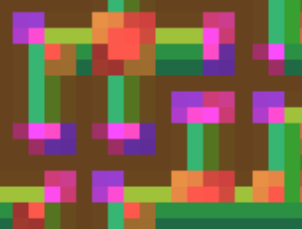}
      \caption{Outer corners on the bottom left break the regularity established by the other inner and outer corners of having 4 class representations.\\}
      \label{fig:cobl-prob}
    \end{subfigure}

    \begin{subfigure}[b]{0.23\textwidth}
      \includegraphics[width=\textwidth]{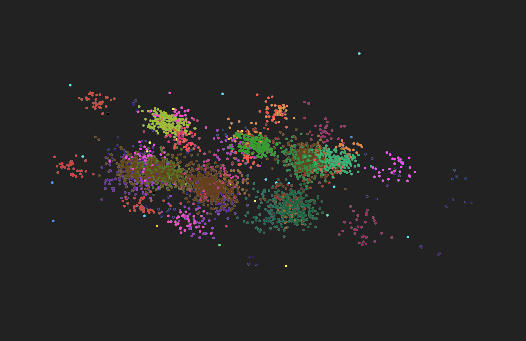}
      \caption{Even with the coloring, it is too cluttered to see what's going on.
      Since the clusters are labeled, we can easily select the four clusters surrounding our interest and hide the rest. 
      }
      \label{fig:cobl-ndsp-all}
    \end{subfigure}
    \hfill
    \begin{subfigure}[b]{0.23\textwidth}
      \includegraphics[width=\textwidth]{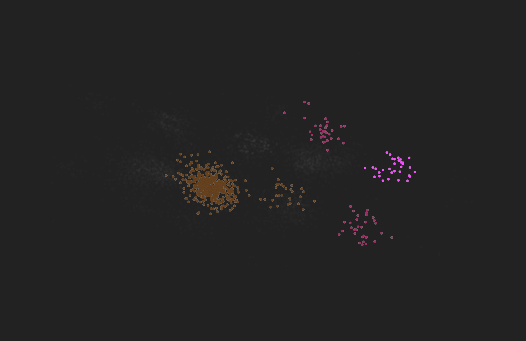}
      \caption{This makes it clear what happened. Basically, the clusters got too close together and got classified as one cluster.
      \\
      }
      \label{fig:cobl-ndsp-focus}
    \end{subfigure}
    
    \begin{subfigure}[b]{0.23\textwidth}
      \includegraphics[width=\textwidth]{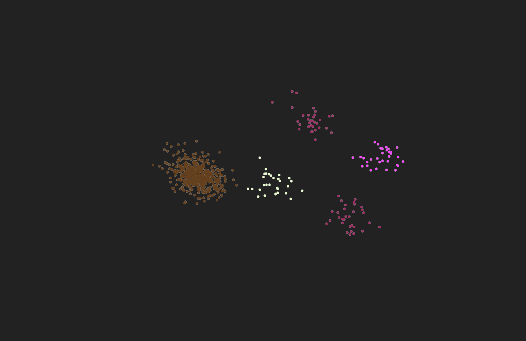}
      \caption{We can easily select the misclassified pixels and add them to a new pixel class.
      \\\\\\\\
      }
      \label{fig:cobl-ndsp-fixed}
    \end{subfigure}
    \hfill
    \begin{subfigure}[b]{0.23\textwidth}
      \includegraphics[width=\textwidth]{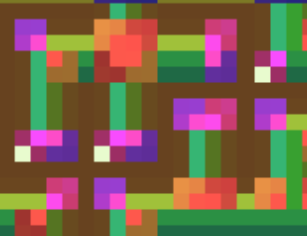}
      \caption{Back in the image view we can confirm that we have corrected the misclassification by seeing that the new pixel class has completed the incomplete bottom left corners.}
      \label{fig:cobl-fixed-vis}
    \end{subfigure}

  %%\hfill
  \caption{Correcting pixel classification issue using NDSP.}
  \label{fig:ndsp-fix-dist}
\end{figure}

We show a case study to illustrate the usefulness of being able to interact with the classes in the pixel distribution. In Fig~\ref{fig:ndsp-fix-dist} an inconsistency is found in the pixel classification that is showing up in PixCol. The classification is exported from PixCol and imported into NDSP, where the clusters are interactively examined. The issue with the classification of the distribution is identified and corrected, and then the modified classification is imported back into PixCol to show that the correction has resolved the inconsistency.

We used these tools and methodologies to examine the output from 3 layers of the policy network: The very first layer: block1.conv, the second conv layer: block1.res1.conv1, and the layer with channel 55 identified in \cite{understanding and controlling}:block2.res1.resadd. Note that the first and second conv layers do have nonlinear, unparameterized layers block1.maxpool, and block1.res1.relu1 between them. These layers do output activations which could be interpreted using this method, however in the interest of time, our focus was directed to conv layers. Also, block2.res1.resadd is about halfway through the network, much deeper than the other two. We briefly discuss our findings in the following sections.

\subsubsection{b1.conv}

We are very impressed with the techniques ability to render the first layers activations interpretable. As could reasonably be predicted, the first layer is involved in basic edge detection tasks, but the exact nature of those edge detections is quite interesting. Each of the classes at this layer can be understood as relating to one of 4 concepts:  the mouse, the cheese, straight paths, or corners.

\begin{figure}[!tbp]

     \centering
    \begin{subfigure}[b]{0.21\textwidth}
      \includegraphics[width=\textwidth]{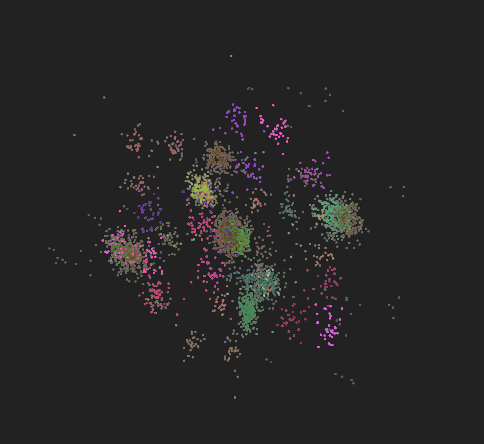}
      \caption{PCA view of corner clusters (pink and purple).}
      \label{fig:pca-corners}
    \end{subfigure}
    \hfill
    \begin{subfigure}[b]{0.21\textwidth}
      \includegraphics[width=\textwidth]{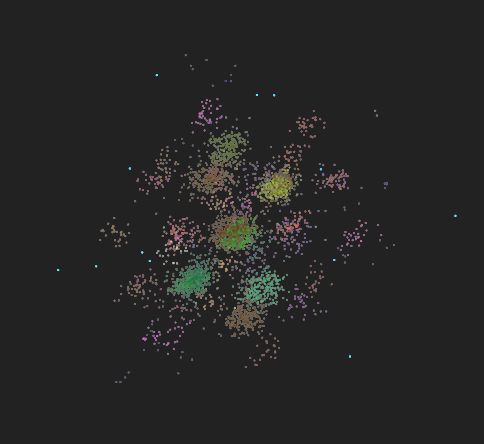}
      \caption{PCA view of mouse pixels (light blue).
      }
      \label{fig:pca-mouse}
    \end{subfigure}
  
  \caption{PCA applied to different points. The is the difference because of how many more corner pixels than mouse pixels there are, or does it represent a difference in how the network conceptualizes?}
  \label{fig:pca_mouse_}
\end{figure}

The mouse is the most interesting concept when examined in NDSP. Unlike the others, it's points do not cluster together, but instead spread apart, moving outside of the cluster of all other classes. This may be caused by the different amounts of pixels, but may also represent differences in how the network will use the concepts. See Fig~\ref{fig:pca_mouse_}.

Because the pixels are so far out, hierarchical clustering assigns them all to separate categories from each other and everything else. K-means clustering on the other hand distributes them among other classes since they are in different directions and are far away from everything. It is only through examining their relation to the previous layer that we could identify them as cheese pixels. Manual clustering does seem to correctly identify them since even though they can be viewed as all around the other points, the view can also be rotated so they all cluster to one side. Further analysis is needed to give a good explanation of the actual structure that this represents.

The cheese is very similar to the mouse, although may not be as extremely dispersed as the mouse is. This may be because of it's relative importance being less than the mouse, since the mouse is the thing that moves, or it could be because the maze we examined had the mouse closer to more other concepts, while the cheese is near relatively fewer. It is totally surrounded by brown floor pixels. Further investigation of other mazes and the surrounding context of the pixels is recommended.

The consistency of the corners is quite impressive. Each type of corner across inner vs outer, top vs bottom, and left vs right has exactly 4 sub distributions. This may be a result of the 3x3 convolution kernel seeing the corner from exactly 4 spots. It may be that these different classes are to be merged into single corner concepts in the next layer, or it could be that the different corners will be extended to combine with other corners and walls into higher level concepts. It is interesting that the clusters in Fig~\ref{fig:cobl-ndsp-fixed} seems to have a geometric relationship with the shape that the pixels found in relation to one another. Further analysis on the pixel distribution of block1.res1.conv1 may clarify which is the case.

The walls and floor are the somewhat inconsistent. part of this seems to be because of aliasing. Some of the walls and floors are thicker or thinner than others. This was not something we had noticed until we started exploring the pixel class locations in PixCol. This seems to cause places where there is or is not a inner wall concept, or just left of wall or right of wall concepts.
Some of the wall classes are quite small and inconsistent. There may be more to these representations than is obvious, although it may also be caused by the noise in the maze texture and wall thickness.

Also worth noting is that the outside edges and corners  of the maze get their own sets of wall and corner classes, consistent along each edge, but not shared between edges or with the rest of the classes inside the maze. This may be because of the way the downstream layers use the outer edges, or, again, it may be an artifact of the convolution process.

Overall, the concepts of layer 1 are not terribly abstract. Basically all of them could be easily implemented manually. However we are curious to know whether the inconsistencies will be abstracted or made use of in deeper layers.

\subsubsection{b1.res1.conv1}

\begin{figure}
    \centering
    \includegraphics[width=1\linewidth]{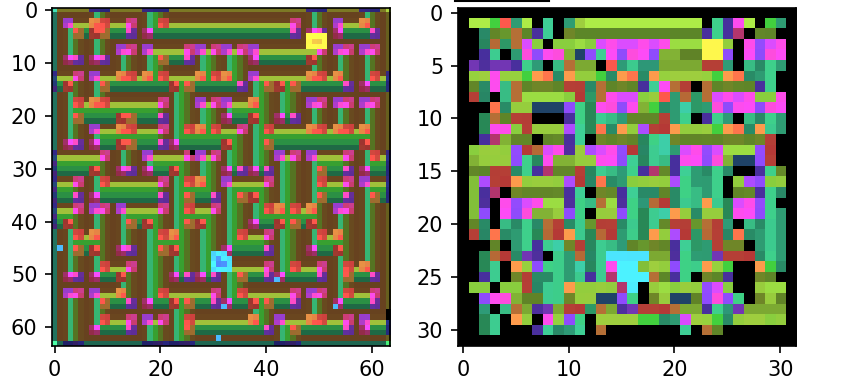}
    \caption{Coloring of block1.conv as reference for coloring block1.res1.conv1}
    \label{fig:PixCol_b1c_b1r1c1}
\end{figure}

The layer b1.res1.conv1 seems to be moving to noticeably higher abstraction concepts. As can be seen in Fig~\ref{fig:PixCol_b1c_b1r1c1}, this seems in part forced by the reduction in number of pixels from the maxpool layer. However, it must be assumed that it is also transforming the representation of the data in a way that moves it closer to a decision on which direction the mouse will move.

The main evidence we can see of this is the expansion of the mouse concept. It appears it may be starting to perform some kind of flood fill algorithm, sending the concept of "mouse goes right" down one path, while sending the concept of "mouse goes up" up another path. There also is minor evidence that the mouse may be altering or resolving maze structure around it as seen by the strange formation of corners to the bottom left of the mouse, and it seeming to push the wall out of the way to the top left. However, the concepts are fuzzy so further investigation is needed.

Some evidence of the higher level of abstraction is seen in the loss of the "floor" and "inner wall" concepts, which seem to have been replaced only by walls. Additionally, the concept of "upper right corner" (purple) seems to have fallen a distance from it's original reference pixels. In some cases falling under a layer of the concept of "upper left corner" while presumably still representing a path between two corners that the agent could travel. It seems that by following rules such as "you may only move from a left wall to a right wall while traveling right, never while traveling left", and "You may only move from a right wall to an upper wall if an appropriate corner is also present" one can still make sense of and navigate this representation of the maze.

This makes the layer more difficult to interpret, but also more rewarding. Unfortunately we do not fully understand the concepts represented on this layer yet, but feel that with more time to cross reference between the PixCol and NDSP, good progress could be made understanding concepts deeper into the network.

\subsubsection{b2.res1.resadd}

As a final experiment with PixCol and NDSP we attempted to look at a much deeper layer, b2.res1.resadd, which is the residual block. This means we are half way through the second of 3 Impala blocks, past 8 conv layers. By this point the image has been reduced to a 16x16 image with 128 channels. The concepts must be much more abstracted and are possibly bleeding together in interesting ways.

We first noticed that hierarchical clustering no longer seems to perform well. The blue shown in Fig~\ref{fig:b2-hi} is all one classification, and every pixel left black was checked and belongs to a classification alone. Clearly the density of the points is quite variable, an idea that is reinforced by examining the pixel distribution in NDSP. Fig~\ref{fig:b2-ndsp} shows the distribution. By looking at the manually colored and K-Means colored versions, it appears that the the mouse concept is pulling more pixels further from the rest of the distribution. The appearance of the colorings suggest there is noise or quite abstracted representations.

Two other possibilities seem worth mentioning, that the network is "throwing out" useless information as it goes, such as deleting the existence of short dead end paths from it's representations of the maze, and so the representation at block2.res1.resadd may not be more abstract, but just contain less unnecessary information. It could also be the case that the  network is making use of orthogonal subspaces, in which case, continuing to represent the output as an image may lose it's sensibility. This is possibly something that could be investigated using extensions to NDSP. Another possible avenue for understanding this layer is to work our way from the shallow layers building up interpretations of the concepts the network is working with at that each layer to inform the interpretation of the next. Indeed, this space seems to have many promising directions for further research.

\begin{figure}[!tbp]

     \centering
    \begin{subfigure}[b]{0.21\textwidth}
      \includegraphics[width=\textwidth]{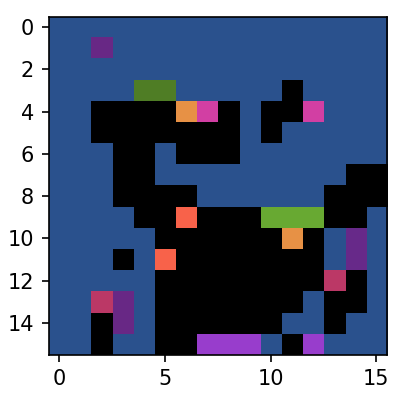}
      \caption{Hierarchical}
      \label{fig:b2-hi}
    \end{subfigure}
    \hfill
    \begin{subfigure}[b]{0.21\textwidth}
      \includegraphics[width=\textwidth]{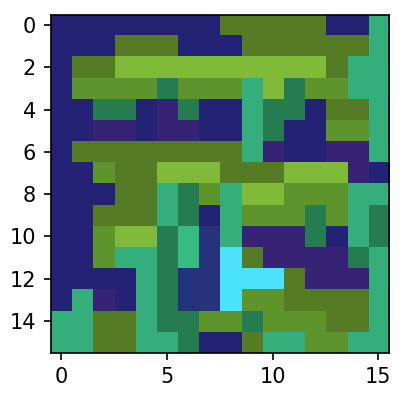}
      \caption{K-Means}
      \label{fig:b2-km}
    \end{subfigure}
      \begin{subfigure}[b]{0.21\textwidth}
      \includegraphics[width=\textwidth]{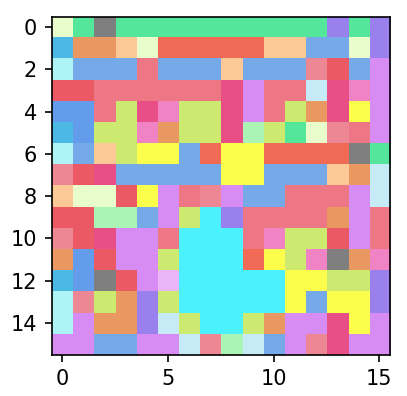}
      \caption{Manual with NDSP}
      \label{fig:b2-ndsp}
    \end{subfigure}
  \caption{PixCol colorings of block2.res1.conv1 pixel classifications by different clustering techniques.}
  \label{fig:b2-pixcol}
\end{figure}

\begin{figure}[!tbp]

     \centering
    \begin{subfigure}[b]{0.21\textwidth}
      \includegraphics[width=\textwidth]{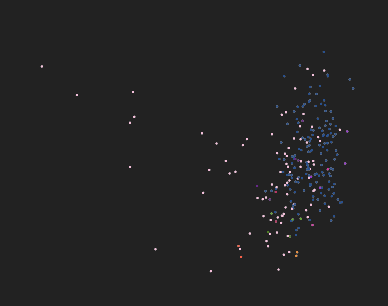}
      \caption{Hierarchical\\(NOTE: the light pink pixels are all single pixel classes, not a single class.)}
      \label{fig:b2-hi}
    \end{subfigure}
    \hfill
    \begin{subfigure}[b]{0.21\textwidth}
      \includegraphics[width=\textwidth]{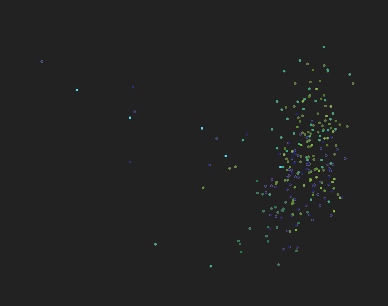}
      \caption{K-Means\vspace{26px}}
      \label{fig:b2-km}
    \end{subfigure}
      \begin{subfigure}[b]{0.21\textwidth}
      \includegraphics[width=\textwidth]{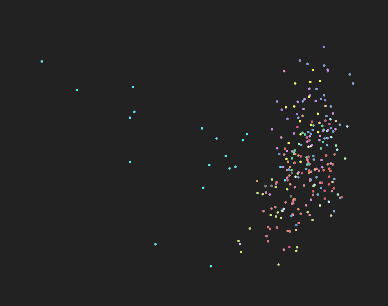}
      \caption{Manual with NDSP}
      \label{fig:b2-ndsp}
    \end{subfigure}
  \caption{NDSP distributions of block2.res1.conv1 pixel classifications by different clustering techniques.}
  \label{fig:b2-ndsp}
\end{figure}

\section{Conclusion}

This study delves into the mechanistic interpretability of reinforcement learning (RL) agents by examining the internal workings of a neural network trained on procedural maze environments. The research uncovers how specific neurons within the network selectively respond to fundamental features such as maze walls and pathways, forming the basis of the model's decision-making process. A significant finding is the identification of goal misgeneralization, where the RL agent develops a bias towards certain navigation strategies, such as consistently moving towards the top right corner, even when explicit goals are absent. Techniques like saliency mapping and feature mapping were instrumental in visualizing these biases, providing critical insights into the network's internal mechanisms and decision-making processes.

The investigation into deeper network layers revealed more abstract and complex representations of the maze environment, necessitating advanced analytical tools for accurate interpretation. Interactive distribution coloring and n-dimensional scatter plots were employed to identify and correct misclassifications within pixel distributions, leading to a nuanced understanding of the network's functionality. This study not only contributes to the field of AI interpretability by offering a comprehensive case study of an RL agent with known misgeneralizations but also highlights the importance of combining multiple interpretability techniques. The findings underscore the need for developing robust and explainable AI systems, particularly in applications demanding high transparency and accountability, and suggest future research directions to mitigate goal misgeneralization and enhance the reliability of RL agents.

\section*{Acknowledgment}
The authors would like to thank our directing professor George Tzanetakis. The authors would like to thank each other. Thanks Triston. Thanks Tristan. 

Thank's also to Alex Turner, Ulisse Mini, Peli Grietzer, Mingwei Li, Carlos Scheidegger for providing guidance understanding and extending their respective projects.

\vspace{12pt}

\end{document}